\documentclass[letterpaper]{article} 
\usepackage{aaai24}
\usepackage{times}  
\usepackage{helvet}  
\usepackage{courier}  
\usepackage[hyphens]{url}  
\usepackage{xcolor}         
\usepackage{graphicx} 
\usepackage{natbib}  
\usepackage{caption} 
\usepackage{algorithm}
\usepackage{algorithmic}
\usepackage{subcaption}
\usepackage{booktabs}
\usepackage{amsmath}

\usepackage{newfloat}
\usepackage{listings}
\DeclareCaptionStyle{ruled}{labelfont=normalfont,labelsep=colon,strut=off} 
\floatstyle{ruled}
\newfloat{listing}{tb}{lst}{}
\floatname{listing}{Listing}
\nocopyright

\makeatletter
\renewcommand*{\@fnsymbol}[1]{\ensuremath{\ifcase#1\or \mathparagraph \else\@ctrerr\fi}}
\makeatother

\title{Understanding the Interplay of Scale, Data, and Bias in Language Models: \\ A Case Study with BERT}

\author{
    Muhammad Ali\thanks{Corresponding author: ali.muh@northeastern.edu}\\
    Swetasudha Panda{\rm *},
    Qinlan Shen{\rm *},
    Michael Wick{\rm *},
    Ari Kobren{\rm *}
}
\affiliations{
    \textsuperscript{\rm \ensuremath{\mathparagraph}}Institute for Experiential AI, Northeastern University\\
    \textsuperscript{\rm *}Oracle Labs
}

\usepackage{bibentry}

\graphicspath{ {./figures/} }

\definecolor{airforceblue}{rgb}{0.36, 0.54, 0.66}

\begin{document}
\pagenumbering{arabic}

\maketitle

\begin{abstract}  
  In the current landscape of language model research, larger models, larger datasets and more compute seems to be the only way to advance towards intelligence. While there have been extensive studies of scaling laws and models' scaling behaviors, the effect of scale on a model's social biases and stereotyping tendencies has received less attention.
  In this study, we explore the influence of model scale and pre-training data on its learnt social biases.
  We focus on BERT---an extremely popular language model---and investigate biases as they show up during language modeling (upstream), as well as during classification applications after fine-tuning (downstream).
  Our experiments on four architecture sizes of BERT demonstrate that pre-training data substantially influences how upstream biases evolve with model scale.
  With increasing scale, models pre-trained on large internet scrapes like Common Crawl exhibit higher toxicity, whereas models pre-trained on moderated data sources like Wikipedia show greater gender stereotypes.
  However, downstream biases generally decrease with increasing model scale, irrespective of the pre-training data. 
  Our results highlight the qualitative role of pre-training data in the biased behavior of language models, an often overlooked aspect in the study of scale. Through a detailed case study of BERT, we shed light on the complex interplay of data and model scale, and investigate how it translates to concrete biases.

  
  
  
\end{abstract}

\section{Introduction}

Large Language Models (LLMs) continue to grow in size at a remarkable rate, with technology companies investing millions in infrastructure to produce ever-larger and more general purpose models.
Modern open weight models like LLaMA, Gemini and Falcon regularly have tens of billions of parameters, showcasing noteworthy capabilities across a range of natural language processing applications.
%


To investigate the performance of LLMs in terms of model parameters, training data size, and compute resources, a rich literature in empirical scaling laws~\citep{hernandez2021scaling,kaplan2020scaling} has emerged, which suggests that bigger is indeed better (in terms of loss).
Recent work on scaling laws has also led to a more comprehensive understanding on the tradeoffs between data size and model parameters with a fixed compute budget~\cite{hoffmann2022training}.
Given the pace of LLM development and the foundational role of scale, studying changes in model behavior with size remains a pressing research problem.


One crucial question that has received less attention, however, is how model scale influences social biases.
LLMs inherently absorb societal biases and harmful stereotypes from data during both pre-training and task-specific fine-tuning.
These biases manifest as \textit{intrinsic} biases within the embedding space, leading to representational harms and stereotyping~\citep{nangia2020crows,nadeem2020stereoset,may-etal-2019-measuring,kurita2019measuring}, and \textit{extrinsic} biases leading to allocative harms~\cite{barocas2017problem} in downstream predictions~\citep{gehman2020realtoxicityprompts,garimella2019women, blodgett2018twitter}. Prior work has shown that pre-trained models can generate toxic language~\cite{gehman2020realtoxicityprompts}, can have disparities in hate speech classification~\cite{sap2019risk}, can perpetuate anti-Muslim bias in text generation~\cite{abid-2021-persistent}, can rely on racial biases even for high stake use cases such as clinical notes~\cite{zhang2020hurtful}, among other failures.


The growing scale of LLMs has been driven, in part, due to their popularity in commercially successful chat applications (e.g. ChatGPT), as well as their instruction following capabilities~\cite{wei2021finetuned}, making them useful for a variety of tasks. Chat and prompting applications tend to use autoregressive, decoder-only Transformer models (e.g. GPT-4, LLaMA, PaLM).
While these models are at the cutting edge, they are often challenging to deploy in many cases, requiring massive compute resources and improvised prompt engineering.
%
In contrast, encoder-decoder (e.g. T5, BART) and encoder-only (e.g. BERT, RoBERTa) Transformer models trained on a Masked Language Modeling (MLM) objective are often lighter, and remain workhorses for NLP applications in industry.
These models continue to be relevant for applications such as summarization, semantic search, sentiment analysis, and a wide array of classification tasks after fine-tuning, as evidenced by their continued success in public machine learning competitions~\cite{pii-detection-removal-from-educational-data,llm-detect-ai-generated-text}.
These models are also affected by biases in the training data similar to autoregressive LLMs, both during pre-training and during the task-specific fine-tuning process.
In this study, we take a step back from the outsize discourse on autoregressive models and focus on encoder-only LLMs (also referred to as MLMs here) due to their widespread use. We present a detailed case study of the influence of scale and data on social biases when pre-training BERT~\cite{devlin2018bert}, a seminal and extremely popular LLM. We focus on one model family specifically to control for variance in model architecture when changing number of parameters.

%


\paragraph{How could scale influence bias?} The prevailing wisdom is that pre-trained LLMs should get more biased as they get bigger. One intuition for this comes directly from the scaling laws themselves.
Models often learn harmful artifacts in the data, and since bigger models fit the data better, biases can aggravate as model size increases.
But there are also  cases  when biases plateau or even decrease with scale ~\citep{srivastava2023beyond}, possibly because scale helps the model learn more reliable task-specific rules without overly relying on shortcut heuristics ~\cite{mccoy19right,bhargava21generalization}. 

Indeed, there are good reasons for why bias should not in general increase with scale, and in fact, should decrease with scale in certain cases.
{\it First}, MLMs such as BERT are known to use shortcut heuristics rather than actually learning more robust heuristics for the downstream task. For example, models fine-tuned for natural language inference (NLI) use the shortcut that when the premise has a high word overlap with the conclusion, the model uses this as a heuristic to predict entailment~\cite{mccoy19right}.
It is likely that something like gender bias is a shortcut heuristic that the model might use to perform a task. And while scale does not solve the problem of shortcut heuristics in NLI, it helps a lot: the jump from the smallest to largest BERT model gets a 23\% absolute increase in accuracy on HANS, and a jump from BERT-base to RoBERTa-base (more data) results in a 19\% improvement \cite{bhargava21generalization}. Similarly, we might hope that scale will help the model learn more reliable task specific rules in favor of biases like gender bias; and so for this reason, a model might actually get less biased with scale.
{\it Second}, many bias and fairness measures, such as equal odds and equal opportunity are functions of the raw statistics that contribute to accuracy, such as false positive rates and false negative rates~\cite{garg2020fairness}. Moreover, downstream allocative harms are frequently measured by these statistics directly. So although scaling laws hold that training loss decreases with scale, indicating that the model might overfit to the bias; at the same time, test accuracy also increases with scale. For this reason, we would expect that fairness measures based on accuracy statistics to actually improve with scale as the accuracy increases. 
{\it Third}, scale isn't the only important factor, the type of data can matter. Different data has different biases. For example, biographies about notable figures on Wikipedia are skewed towards men over women~\cite{tripodi2023ms}, and large web scrapes like the common-crawl are likely more diverse in overall topics covered, but also much more toxic.

\paragraph{Our contributions.} We study MLMs to see how social biases evolve as we vary model scale. 
We pre-train four architecture sizes of BERT (\texttt{mini}, \texttt{small}, \texttt{medium} and \texttt{base}, i.e., up to 110M parameters).
A core focus of our study is the pre-training dataset.
We experiment with English Wikipedia and the CC-100 English subset of Common Crawl. 
For each model size and type of training data,
we compute biases in the representations upstream and performance disparities downstream, leveraging specific measures of language biases from prior research~\cite{steed2022upstream}.
We measure upstream bias along two dimensions: disparities in gender pronoun probabilities and sentiment of generated text; downstream impact is measured by fine-tuning models on a toxicity classification task, where we evaluate differences is false positive rates across a diverse set of demographic groups from prior work~\cite{dixon2018measuring} (e.g. {\it Muslim}, {\it gay} etc.)


Our findings underscore the crucial role of pre-training data
as models increase in size. 
For models pre-trained with CC-100, upstream biases generally increase with model size.
Conversely, models pre-trained on Wikipedia show greater gender stereotyping as models increase in size. In both cases, we find that downstream biases decrease with increasing model size. 
However, models consistently associate certain identities such as \textit{gay} and \textit{homosexual} with toxicity, independent of parameter size or type of pre-training data.
This finding aligns with prior work ~\citet{steed2022upstream,panda-etal-2022-dont}, that downstream biases are largely influenced by biased artifacts in the fine-tuning dataset, and not the pre-training data. 

We then inspect the datasets themselves to identify the reasons of observed biases, and find that indeed CC-100 contains more negative sentiment towards our measured identity groups compared to Wikipedia, which is picked up by larger models. We also find evidence of Wikipedia encoding more male pronoun co-occurences in articles related to occupations, which might explain the increased gender disparities in models pre-trained on Wikipedia.

In summary, we conduct a detailed case study through BERT---an extremely popular model---and investigate the impact of pre-training data, model scale, and observed social biases, both in terms of the masked language modeling task, as well as in downstream classification settings.

\section{Related Work}
Our study broadly relates to three strands in the literature: fairness and auditing of machine learning algorithms in general, the study of biases in natural language processing more specifically, and scaling laws for large language models.

First, a rich literature in computer science investigates issues of fairness in machine learning~\cite{dwork2012fairness,barocas2023fairness}, measuring the different types of harms these systems can have---such as denigration, stereotyping, differential quality of service etc.~\cite{barocas2017problem,weerts2021introduction}.
Notably, prior work has documented racial and gender disparities for commercial gender classification systems~\cite{buolamwini2018gender}, racial disparities in criminal recidivism prediction~\cite{angwin2022machine}, gender disparities in the delivery of job advertising~\cite{datta2014automated,ali2019discrimination}, among others. Many of these studies only rely on ``black box'' access to machine learning systems, and have to conduct clever {\it audits} to measure disparate outcomes~\cite{metaxa2021auditing}. Our work is connected to this literature in its goal of measuring inadvertent harms of a machine learning system, albeit with ``white box'' access to the model's output probabilities and weights.

Within natural language processing (NLP) specifically, prior work has also discussed biased and disparate outcomes for users. \citet{blodgett2016demographic} was one of the earliest works documenting racial disparities, showing how dependency parsing tools struggle on text for African American English on Twitter. Similarly, \citet{caliskan2017semantics} demonstrated how word embeddings learnt from text corpora can contain gender biases. In the context of large language models---which power most of modern NLP---recent work has documented negative associations for people with disabilities~\cite{hutchinson2020social}, anti-Muslim bias~\cite{abid-2021-persistent}, and a general propensity to generate toxic text~\cite{gehman2020realtoxicityprompts}.
There have also been efforts to construct benchmarks that can yield repeatable measurements of bias across many different language models. This includes benchmarks such as WinoBias for coreference resolution~\cite{zhao2018gender}, BBQ and UNQOVER for question-answering~\cite{parrish2021bbq,li2020unqovering}, BBNLI for natural language inference~\cite{baldini2023keeping}, StereoSet for measuring stereotypical associations~\cite{nadeem2020stereoset}, among others. Further, large benchmarking efforts such as BIG-bench~\cite{srivastava2023beyond} have been able to provide insights into the relationship between model scale and performance on bias benchmarks, which is one of the objectives of our study. We now know from ~\citet{srivastava2023beyond} that for auto-regressive models, bias (as measured via UNQOVER, BBQ etc.) typically increases in ambiguous prompts, and that it can decrease for narrow, unambiguous prompts. We similarly study the relation of bias with scale, but in the context of MLM models, which have distinct upstream and downstream applications, and with an added focus on the pre-training dataset used. Closest to our study is \citet{steed2022upstream}'s work on upstream and downstream biases for MLMs, in which they investigate the {\it bias transfer hypothesis}---can upstream debiasing methods improve disparities in downstream performance? They find that upstream mitigation does little to address downstream biases, and that downstream disparities are better explained by biases in the fine-tuning data.


In parallel, empirical scaling laws related to LLM performance have been the subject of extensive investigation in recent research~\cite{hestness2017deep,kaplan2020scaling}. These studies have found a power-law scaling relationships with model size, dataset size, and computational resources, i.e. an increase in either almost always leads to a decrease in loss. Recently, \citet{hoffmann2022training} have also led to a clearer understanding of the tradeoffs between training data size (number of tokens) and model parameters, yielding a unified formula for compute-optimal training, which has already been applied to specific model settings \citep{clark2022unified,gordon2021data,henighan2020scaling,tay2022scaling}. However, it is noteworthy that the scalability of LLMs does not universally translate to improved performance across all downstream tasks, as demonstrated by~\citet{ganguli2022predictability}. Similarly, recent work by \citet{wei2022emergent} highlights emergent abilities unique to larger models not predicted by traditional scaling laws. In response to work in scaling laws, there has also been pushback from critics. Notably, \citet{bender2021dangers} highlighted the rising environmental and financial costs of model pre-training, and the lack of diversity in training data. Closely related to our study, \citet{birhane2023hate} study scaling laws in the context of hateful content present in the LAION family of datasets, popularly used to pre-train text to image diffusion models. They find that as data scale increases, the tendency of models to associate Black faces with categories like ``criminal'' can significantly increase.

Our work lies at the intersection of these research threads. We contribute to the ongoing practice of measuring the adverse outcomes of machine learning systems. Our work also contributes to ongoing work on scaling laws, with a specific focus on bias, and how it is picked up from pre-training data. 

\section{Methods}
\label{sec:methods}
In this section, we cover the models we train, our training configuration, the datasets used to train these models, and the metrics we use to measure bias.

\subsection{Models}
\label{subsec:models}
We 
experiment with four architecture sizes of BERT: BERT-Mini, BERT-Small, BERT-Medium, BERT-Base. 
While originally introduced in the context of model distillation~\cite{turc2019well}, we find that these models provide a good testbed for experimenting with model scale, while holding the architecture constant.
Table~\ref{tab:model_params} shows the number of layers, hidden embedding size, and the number of parameters in each case. 
Following \cite{turc2019well}, we fix the number of attention heads to $H/64$, where $H$ is the hidden embedding size. We use the publicly available architecture implementations of miniature BERT architectures via HuggingFace\footnote{BERT-Medium, e.g. \url{https://huggingface.co/google/bert_uncased_L-8_H-512_A-8}}.

\begin{table}[h]
    \centering
    \begin{tabular}{lccc}        
        \toprule
        {\bf Model} & $\mathbf{L}$ & $\mathbf{H}$ & {\bf Parameters}\\
        \midrule
         BERT-Mini & 4 & 256 & 11.3M \\
         BERT-Small & 4 & 512 & 29.1M \\
         BERT-Medium & 8 & 512 & 41.7M \\
         BERT-Base & 12 & 768 & 110.1M \\
    \end{tabular}
    \caption{BERT architecture specifications for our models. We vary number of layers ($L$) and hidden embedding size ($H$).}
    \label{tab:model_params}
\end{table}

\subsection{Pre-Training Data}
For each model size, we pre-train on three different datasets, on a masked language modeling objective: (a) CC-100-EN: English subset of Common Crawl \citep{conneau2019unsupervised}, (b) English Wikipedia, and (c) a combination of CC-100-EN and Wikipedia in a multi-task setup.
Text on Wikipedia data is curated by a set of editors, and often goes through moderation, quality control and edits. 
Common Crawl~\cite{wenzek2019ccnet}, on the other hand, is a massive unconstrained crawl of the internet and therefore, is very likely to include stereotypes as well as toxic and abusive statements \citep{luccioni2021s,gehman2020realtoxicityprompts}.
Our choice of these datasets for pre-training is based on these content differences, such that we can contrast language biases after model pre-training.

\paragraph{Pre-training configuration.} For each model size in Table~\ref{tab:model_params}, we pre-train the model for $8,000$ training steps on the chosen pre-training data. We seed our data shuffling consistently to make sure that each model gets exposed to the same set of tokens from the data, which prior work has shown to be fundamental in models' scaling behavior~\cite{kaplan2020scaling,hoffmann2022training}.
When combining datasets, we run combined training with the two datasets interleaved, i.e., one update step on CC-100-EN, followed by one update step on Wikipedia.
As a result, each mini-batch consists only of data from one of the pre-training datasets, and Wikipedia is up-sampled relative to CC-100-EN.

\subsection{Metrics} \label{subsec:bias}
We use a series of metrics from prior work to measure biases at different points. First, we evaluate biases intrinsic to the model itself, i.e. relating to the masked language modeling task it's trained on. Second, we fine-tune each model for a downstream classification task and evaluate how its scale and pre-training data affects false positive rates across demographic groups. Third, we use linguistic analysis on the pre-training datasets themselves to understand the provenance of our observed biases. Here, we describe each of these metrics in detail.

\paragraph{Upstream bias metrics.} We use two metrics from prior work~\cite{steed2022upstream} to evaluate upstream biases in our pre-trained models.

{\it First}, we evaluate gender bias using an extension of log probability bias score from \citet{kurita2019measuring}. We specifically use the version of this metric used in \citet{steed2022upstream}, where templates are constructed for a list of 28 professions taken from the {\it Bias in Bios} dataset~\cite{deart2019bias}. The original dataset is built from Common Crawl, which includes over 400,000 online biographies from 28 occupations.
The dataset does not include self-reported gender; we refer to the pronouns in each biography to denote gender. In our use-case, for the list of 28 professions, we use templates of the form \texttt{\{pronoun\} is a(n) \{occupation\}} to measure the model's propensity towards either he/him or she/her pronouns. To increase the robustness of our measurements, we also include template variations from~\citet{bartl2020unmasking}, e.g. \texttt{\{pronoun\} applied to the position of \{occupation\}}. For each occupation $y$ and pronoun $g$, we compute the model's probability $p_{y,g}$ for the template. To control for baseline differences for pronouns, we also compute prior probability $\pi_{y,g}$ for a template where only the pronoun is present but the occupation is masked, e.g., \texttt{he is a [MASK]}. We define our upstream gender bias metric as the difference in these probabilities: 
\begin{equation}
\log \frac{p_{y,\text{she/her}}}{\pi_{y,\text{she/her}}} - \log\frac{p_{y,\text{he/him}}}{\pi_{y,\text{he/him}}}
\label{eq:log_prob_bias}
\end{equation}

\noindent A higher absolute probability gap suggests that a model associates one gender much more with an occupation, while a value close to zero implies equal association during masked language modeling.

{\it Second}, we evaluate upstream biases beyond gender, and for a diverse set of demographic groups. Following \citet{hutchinson2020social}, we rely on sentiment analysis to measure upstream bias. Again, we re-use methodology from \citet{steed2022upstream} and construct templates of the form \texttt{\{identity\} \{person\} is [MASK]}. The \texttt{\{identity\}} term consists of about $50$ diverse identity groups such ``Muslim'', ``Jewish'', ``elderly'', ``gay'' etc., taken from \citet{dixon2018measuring}. The original dataset consists of (a) 130,000 public comments from Wikipedia Talk pages, annotated for toxicity, which mention these identity groups; and (b) a synthetic test set to evaluate disparities in toxicity classification. We leverage the identity groups to generate templates for upstream biases, and the synthetic test set to evaluate downstream biases later.
The \texttt{\{person\}} part of the template includes phrases like ``people'', ``spouse'' etc. to increase the number of templates we measure. We compute the 20 most likely tokens for \texttt{[MASK]} for each template.
We then use a pre-trained RoBERTa~\cite{liu2019roberta} sentiment classifier\footnote{\label{hf_model}\url{https://huggingface.co/cardiffnlp/twitter-roberta-base-sentiment}} trained on the TweetEval benchmark~\cite{barbieri2020tweeteval} to measure the average {\it negative} sentiment for each identity group's completed prompts. We focus on negative sentiment in particular as a proxy for toxicity and negative associations similar to prior work~\cite{steed2022upstream, hutchinson2020social}, and due to its potential of introducing representational harms~\cite{barocas2017problem}.


\paragraph{Downstream bias metrics.} To evaluate downstream biases, we fine-tune our pre-trained model on a toxicity classification task, and compare false positive rates (FPR) across different identity groups. The FPR of a group $g$ in the data is defined as $$FPR_g = \frac{FP_g}{FP_g + TN_g} = \frac{FP_g}{N_g}$$

\noindent Here, $FP_g$ indicates the false positives in classification, $TN_g$ is true negatives, and $N_g$ are total number of ground truth negatives (i.e. non-toxic sentences), all for group $g$ specifically. We focus on false positives since they can result in concrete allocative harms such as over-moderation and de-platforming \cite{jhaver2021evaluating} if such classifiers were to be used for toxicity classification. Further, prior work \cite{steed2022upstream} has successfully used FPR to quantify downstream performance disparities.
We use the synthetic test set from \citet{dixon2018measuring} as our toxicity classification task; the dataset contains $89$K examples created using templates of both toxic and non-toxic phrases which are filled in with the $50$ identity terms we also use in our upstream bias measurement.
Following the original paper, we divide the synthetic dataset into 75$\%$ training and 25$\%$ (split equally into validation and test).
To build a classifier, we attach a sequence classification head to our pre-trained model and fine-tune for 3 epochs.

\paragraph{Dataset bias metrics.} To investigate the provenance of our observed biases, we measure biases within the pre-training datasets themselves.

{\it First}, to compare gender associations, we analyze differences in \texttt{(pronoun, occupation)} pair co-occurrences between the two pre-training corpora using weighted log odds ratio with a Dirichlet prior~\cite{monroe2008fightin}. Log odds ratio is an alternate to tf-idf and similar word score methods to compare word importance across documents or corpora. In its simplest form, the log odds of word $w$ in a corpus $i$ where it occurs with frequency $f_w^i$ is defined as $log \; O_w^i = log \; \frac{f_w^i}{1 - f_w^i}$; the {\it log odds ratio} can then be used to compare word importance between corpus $i$ and $j$ as:

\begin{equation}
    \label{eq:log_odds}
    log\; \frac{O_w^i}{O_w^j} = log \; \frac{f_w^i}{1 - f_w^i} - log \; \frac{f_w^j}{1 - f_w^j}
\end{equation}

\noindent We use a model-based variant of this measure that is more robust to low frequencies, specifically the weighted log odds ratio with a Dirichlet prior; we refer the reader to \citet{monroe2008fightin} for a more detailed discussion.

{\it Second}, we re-use the sentiment model\textsuperscript{\ref{hf_model}} used for upstream bias to compare sentiment in the pre-training datasets themselves. For both CC-100-EN and Wikipedia, we extract sentences that mention our list of identity groups. We then use the sentiment model to compute average negative sentiment across all sentences that mention a group.

\section{Results}
After our pre-training process, we obtain three variants (Wikipedia, CC-100-EN, and Wikipedia + CC-100-EN) of each model size (Mini, Small, Medium, Base), i.e. twelve pre-trained LLMs in total. We first measure upstream biases in all these models using our two metrics; second, we fine-tune each model to the downstream task of toxicity classification to measure downstream biases. Finally, we use our dataset bias metrics to measure the pre-training dataset themselves, and investigate the provenance of the biases we observe. Here, we present our results from these experiments.







\subsection{Upstream biases can increase with model size}
\label{subsec:upstream}

\begin{figure*}[th]
\centering
\begin{subfigure}{.48\textwidth}
  \centering
  \includegraphics[width=\linewidth]{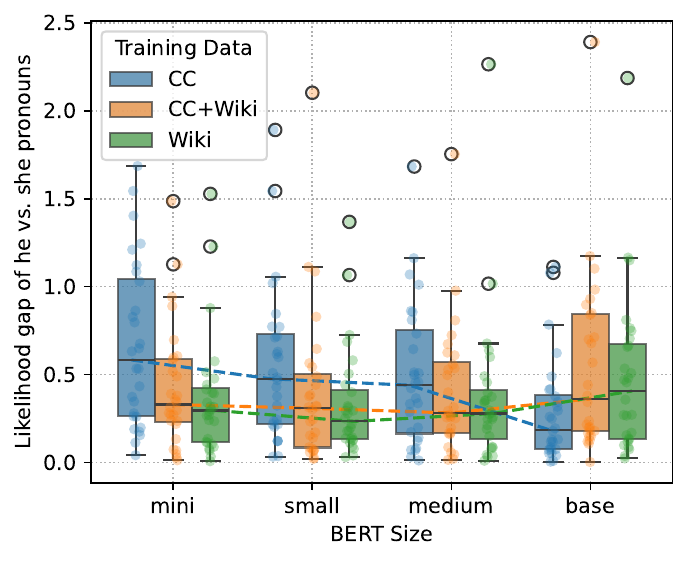}
  \caption{Upstream bias: gender stereotyping}
  \label{fig:upstream_sentiment_occupation}
\end{subfigure}
\begin{subfigure}{.48\textwidth}  
  \centering
  \includegraphics[width=\linewidth]{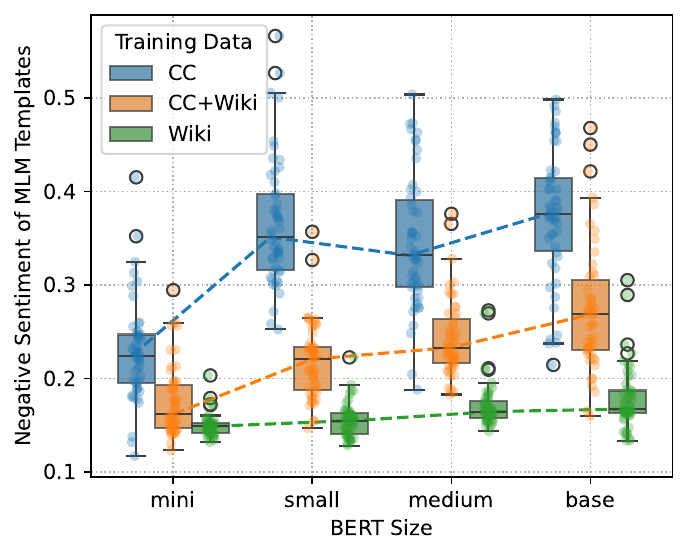}
  \caption{Upstream bias: negative sentiment}
  \label{fig:upstream_sentiment_toxicity}
\end{subfigure}
\caption{Upstream biases for each model size and pre-training data type in terms of our bias metrics: (a) log probability gaps between \texttt{he/him} and \texttt{she/her} pronouns for prompts related to occupations (b) average negative sentiment for masked language modeling completions related to multiple identity groups.}
\label{fig:upstream}
\end{figure*}

We begin by evaluating gender bias upstream using our implementation log probability bias score (Equation~\ref{eq:log_prob_bias}). Figure~\ref{fig:upstream_sentiment_occupation} shows absolute log probability gap between \texttt{he/him} and \texttt{she/her} pronouns for prompts related to 28 occupations, for all 12 of our models. Since we use multiple occupations for our metric, we visualize the probability gap as a distribution across these occupations. Higher values suggest a skew towards either pronoun, while lower values suggest equal likelihood, and therefore better gender representation. We observe that for models pre-trained on Wikipedia (green), gender stereotypes slightly increase with model size---as seen in the increased variance and median. However, for models pre-trained on CC-100-EN (blue), gender stereotypes seemingly decrease with model size.
For models pre-trained on the combination (orange), we do not observe a consistent trend across model sizes.
We qualitatively observe that occupations such as ``nurse'', ``yoga teacher'' and ``software engineer'' consistently appear as outliers across model types.



%


We then measure upstream bias with our second metric, which is the average negative sentiment for prompt completions that relate to 50 identity groups. For each of our pre-trained models, we compute the average negative sentiment for multiple MLM prompts relating to each identity---Figure~\ref{fig:upstream_sentiment_toxicity} shows the distribution of these negative sentiment scores.
Here, we note that as model size increases, we observe a general upward trend in upstream bias, regardless of pre-training dataset. 
We also observe that models pre-trained on CC-100-EN achieve the highest average negative sentiment scores, followed by models pre-trained on the combination of CC-100-EN and Wikipedia. Models pre-trained on Wikipedia exhibit comparatively the lowest average negative sentiment in our experiments. 
%
Qualitatively, in case of models pre-trained on CC-100-EN, we notice frequent abusive mentions (e.g., ``stupid'',  ``sick'', ``insane'') on the list of top words predicted by the model. We also find evidence of these models generating (unfortunate) sentences such as ``Muslim people are dangerous''.
Identities such as ``elderly'', ``deaf'' and ``Muslim'' are the most frequent outliers across model sizes, which aligns with prior work~\cite{dixon2018measuring, abid-2021-persistent}.
In contrast, for models pre-trained on Wikipedia, we note MLM completions associated with lower negative sentiment such as ``wrong'', ``injured'', ``wounded'' etc.
These differences illustrate the effect of both model scale and training data on output toxicity, suggesting that larger models are more capable of learning biases from the data---particularly when that data is from an unmoderated source.



\begin{figure}[h]
\centering
\includegraphics[width=\linewidth]{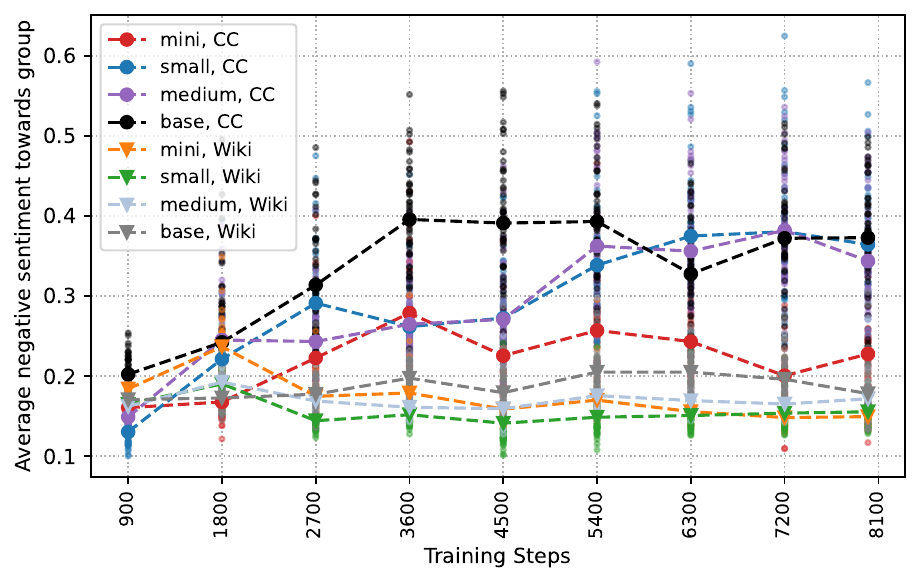}
\caption{Upstream biases (measured via negative sentiment associations) over the course of pre-training. Models pre-trained on CC-100 generally result in higher bias scores compared to models pre-trained on Wikipedia.}
\label{fig:upstream_sentiment_timeline}
\end{figure}

\paragraph{Evolution of bias over the training process.} We also monitor the change in upstream bias (measured via sentiment) during the course of the pre-training process. We checkpoint all models after every 900 training steps during the training process, and compute negative sentiment for each identity group at these checkpoints. Figure~\ref{fig:upstream_sentiment_timeline} shows how upstream biases grow over time in our experiments. Each small point shows the average negative sentiment for an identity group, the large points connected via lines show the average of average negative sentiment for each model.
Similar to our final measurement in Figure~\ref{fig:upstream_sentiment_toxicity}, we notice here too that models trained on CC-100-EN (except BERT-Mini) have higher upstream bias. Models trained on Wikipedia consistently have lower upstream bias and interestingly this does not increase or vary over training.


\begin{figure*}[ht]
\centering
\includegraphics[width=0.75\textwidth]{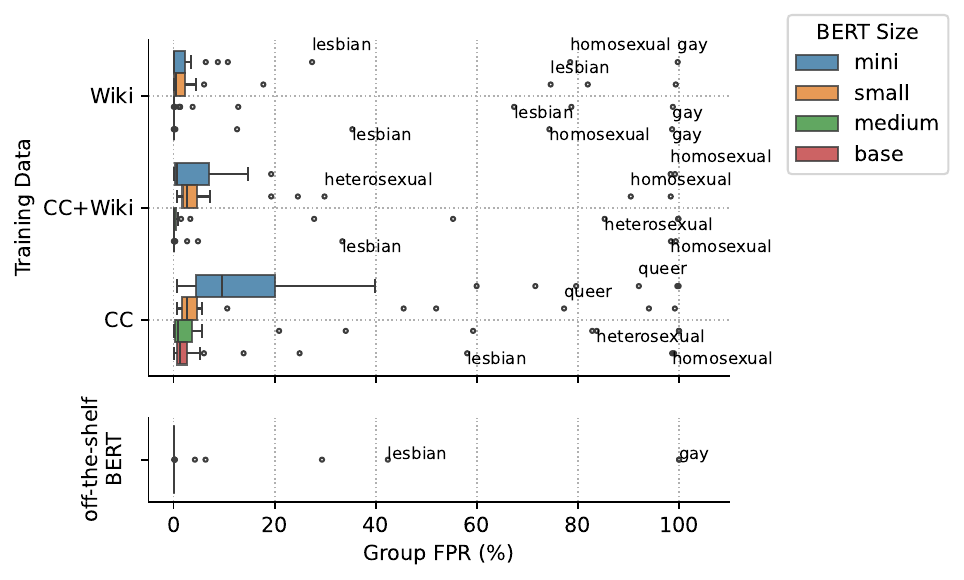}
\caption{Downstream biases evaluated on toxicity classification data from \citet{dixon2018measuring}. For each model size and type of pre-training data, false positive rates (FPR) for each identity group are shown. Median FPR and variance of FPRs decreases as models grow larger, but some outliers remain.}
\label{fig:downstream_fpr_summary}
\end{figure*}

\subsection{Larger models make more robust downstream classifiers}
\label{subsec:downstream}

Next, we turn our attention to downstream biases of our pre-trained models. We attach a classification head to each of our models and fine-tune them for the task of toxicity classification. Following prior work~\cite{steed2022upstream, panda-etal-2022-dont}, we use the synthetic toxicity classification data from \citet{dixon2018measuring} for this task. We also evaluate an off-the-shelf BERT (\texttt{bert-base-uncased}) from HuggingFace.
%
As described earlier, we evaluate downstream biases in terms of differences in false positive rate (FPR) for sentences relating to each identity group.
A higher FPR for a group indicates higher downstream biases, since the model is more likely to falsely flag mentions of that group as toxic, potentially leading to discriminatory censorship.
Ideally, we would like a model to have low FPR on each identity \textit{and} low variance in FPR across all groups.

Figure~\ref{fig:downstream_fpr_summary} shows a distribution of FPR for each model size and pre-training dataset after fine-tuning. We observe that the median FPR decreases as model size increases, regardless of pre-training data; similarly, we note that the variance of FPR across groups decreases for larger models as well. This suggests that after fine-tuning, larger models make more robust classifiers---they make fewer false positive errors for all groups in our experiments.

The decrease in downstream biases with scale can have a few explanations.
First, MLMs  are known to use shortcut heuristics instead of  task-specific robust heuristics  ( e.g., model fine-tuned for Natural Language Inference (NLI) uses  high word overlap with the conclusion, to predict entailment ~\cite{mccoy19right}; scale improves these results \cite{bhargava21generalization}). The model might latch on to certain identities as shortcuts to predict toxicity. 
Second, allocative harms are frequently measured directly using accuracy statistics e.g., FPR in our case. As scaling laws suggest that test accuracy increases with scale, downstream bias statistics will improve as well. 
%
%

%

However, certain identity groups such as ``gay'', ``queer'' and ``homosexual'' consistently show up as outliers in terms of FPR, regardless of model size or type of pre-training data. Prior work~\cite{steed2022upstream} has shown that downstream disparities are largely explained by the fine-tuning data; our observed outliers likely appear disproportionately in toxic sentences, leading to a higher FPR.
This suggests that while larger versions of BERT make more robust downstream classifiers, they are not able to address biases against extreme outliers.


\subsection{Associations in pre-training data influence bias}
\label{subsec:data}

%
%
Finally, we investigate the impact of pre-training data on the biases we observe.
While downstream biases can be explained as an artifact of the fine-tuning data~\cite{steed2022upstream}, we suspect a much tighter coupling between upstream biases and choice of pre-training data.

To understand our observed upstream gender biases (Figure~\ref{fig:upstream_sentiment_occupation}), we use weighted log odds with a Dirichlet prior~\citep{monroe2008fightin} and compare \texttt{(pronoun, occupation)} pair occurrences between CC-100-EN and Wikipedia. Specifically (in terms of Equation~\ref{eq:log_odds}), for each occupation $o \in \{\text{journalist, physician, painter},...\}$, and pronoun $p \in \{ \{\text{he, him, his, himself}\}, \{\text{she, her, hers, herself}\}\}$ we measure: $$log \; \frac{O^{\text{CC-100}}_{o, p}}{O^{\text{Wiki}}_{o, p}}$$

\noindent A positive value indicates a co-occurrence is more likely in CC-100-EN than in Wikipedia, while a negative value means it is more likely in Wikipedia. Also note that we count frequencies for a set of pronouns and not singular pronouns for more robust counting. Further, to normalize for variance, we z-normalize the log odds; using the one-sided critical value for $p=0.05$, we only consider $z > 1.645$ to be a significant difference between both datasets. Table~\ref{tab:data_logodds_differences} shows the 10 occupations with the highest weighted log odds between CC-100-EN and Wikipedia.

\begin{table}[t]
    \centering
    \begin{tabular}{lll}
        {\bf Occupation} & \multicolumn{2}{c}{\bf Pronouns}\\
        & \multicolumn{1}{c}{\bf M} & \multicolumn{1}{c}{\bf F}\\
        \midrule
        teacher & {\bf -3.37*} & {\bf 2.47*} \\
        professor & {\bf -4.14*} & 1.33 \\
        nurse & -0.25 & {\bf 2.87*} \\
        model & -1.00 & {\bf -1.64*} \\
        journalist & -0.31 & -1.06 \\
        painter & -1.26 & -0.06 \\
        physician & -1.16 & -0.16 \\
        composer & -1.19 & -0.11 \\
        attorney & -0.56 & 0.65 \\
        photographer & -0.53 & 0.64 \\
    \end{tabular}
    \caption{Weighted log odds (z-normalized) for occupation, pronoun pairs between CC-100-EN and Wikipedia. $\mathbf{M} = \{\text{he, him, his, himself}\}$, $\mathbf{F} = \{\text{she, her, hers, herself}\}$. Positive values indicate skew towards CC-100-EN, negative values indicate skew towards Wikipedia; $p < 0.05$ shown in {\bf bold}.}
    \label{tab:data_logodds_differences}
\end{table}

While we observe many differences that are not significant, for certain occupations, Wikipedia indeed encodes greater gender stereotypes, e.g., ``professor'' has significantly higher masculine pronoun associations, and ``model'' has higher feminine pronoun associations. Interestingly, ``teacher'' is the only occupation that has significant stereotypical associations in both datasets: masculine in Wikipedia, feminine in CC-100-EN.
%
One trend, despite lack of significance, is that co-occurrences with masculine pronouns are overall more common in Wikipedia than CC-100 (larger negative values in {\bf M} column).
This may be reflective of a broader trend on Wikipedia, beyond gendered stereotypes for specific occupations, where the vast majority of biographical articles are about men, due to biases in who is perceived as notable ~\citep{tripodi2023ms}. Conversely, while large web scrapes like CC-100 are more diverse in overall topics covered, these might involve more toxic text.

\begin{figure}
  \centering
  \includegraphics[width=\linewidth]{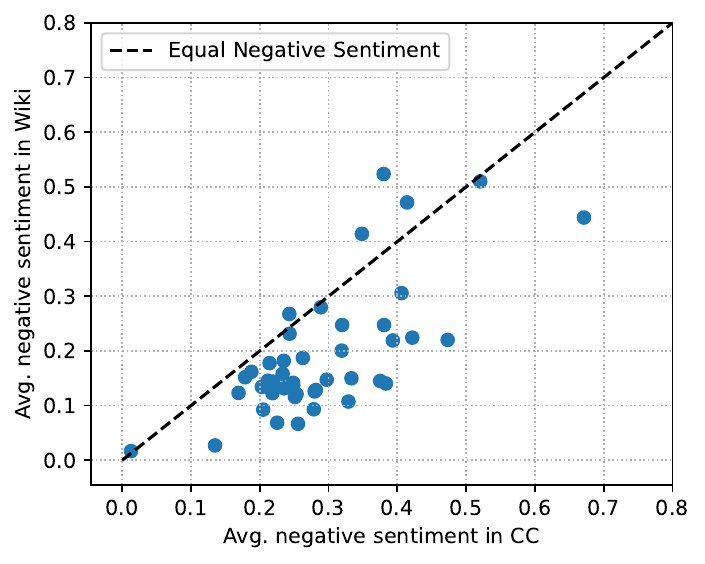}
  \caption{Average negative sentiment for sentences in pre-training data that mention our studied identity groups. CC-100-EN (x-axis) almost always encodes more negative sentiment.}
  \label{fig:data_sentiment_scatterplot}
\end{figure}

To understand the provenance of our upstream sentiment biases (Figure~\ref{fig:upstream_sentiment_toxicity}), we extract sentences from both CC-100-EN and Wikipedia that mention our list of identity groups, and re-use our sentiment classifier to measure average negative sentiment. This allows us to compare whether one dataset a priori encodes more negative sentiment towards a group, which could be picked up by a model during training.
Figure~\ref{fig:data_sentiment_scatterplot} shows average negative sentiment for each identity group in both datasets.
We find that CC-100-EN shows higher negative sentiment for most identity groups compared to Wikipedia. In fact, with the exception of a few outliers, CC-100-EN consistently has more negative sentiment.
We suspect that our observed upstream biases are an artifact of this difference, and that larger models do a better job of picking up these aspects of the data (e.g., Figure~\ref{fig:upstream_sentiment_timeline}).



\section{Limitations}
Our approach has multiple limitations. First, while we limit our analysis to a single model family to reduce variance in architecture, it limits the ecological validity of our results. Our results therefore cannot be generalized to all modern LLM architectures, and instead provide a detailed look into BERT specifically. Second, compared to state-of-the-art compute intensive training procedures, our pre-training process is quite rudimentary. We limit the training process to only 8000 training steps as a heuristic to upper bound the amount of compute that each model uses; this is a simplification and does not lead to a model as powerful as those available through model hubs like HuggingFace. Third, while we make sure to evaluate bias holistically by examining both upstream and downstream differences, our bias metrics---such as log probability gaps and sentiment---are not definitive ways of measuring {\it bias}. Bias is a complex, socio-technical, and sometimes ill-defined notion whose meaning can vary across domains and tasks. While we rely on metrics from prior work, our measures are prone to the same pitfalls and limitations in validity that most bias measurement work in NLP suffers from~\cite{goldfarb2023prompt}.

\section{Concluding Discussion}
Our study provides a detailed case study on the interplay of scale, pre-training data, and bias with a specific focus on BERT, a widely used LLM. We find evidence that larger models are able to encode more biases upstream. Importantly, we observe that larger models, combined with unmoderated data, can lead to worse results for the task of masked language modeling. However, larger models can also produce more robust downstream classifiers after fine-tuning.

While MLMs like BERT do not represent the state-of-the-art in the rapidly developing landscape of LLM research, they remain extremely relevant for several applied natural language processing problems. Our investigation of bias is particularly relevant to practitioners who fine-tune embedding models for their tasks. In these applied use-cases, our results shed light on how scale and training data together can lead to different kinds of biases. We encourage practitioners to be aware of the biases their training datasets can introduce, and to actively measure these artifacts during the development process. On a more general level, our study highlights the role that training data can play in scaling, especially as it relates to biased model behavior. Our results also suggest that mixing in a moderated, high quality data source (e.g., Wikipedia) with larger datasets (e.g., CC-100, The Pile~\cite{gao2020pile}) might be an approach to alleviate biases---we leave a full exploration of this direction to future work.

Our analyses also underscore the limitations that exist in metrics used to measure {\it bias}, which is a nuanced socio-technical concept, whose meaning changes across tasks and domains. Negative sentiment and gaps in gender representation---as used here---are well-scoped ways of expressing bias that can be useful for different domains. Negative sentiment, for instance, could be a useful measure of bias for LLM use in chatbots or auto-complete tools; differences in gender likelihood could be useful for measuring bias in resumé or search ranking, but they are not universal measures of linguistic bias. As seen in our results, depending on the choice of bias metric, a measurement of model behavior can look quite different. This aligns with prior work \cite{goldfarb2023prompt, blodgett2021stereotyping} which shows that measuring bias or fairness can be a challenging undertaking, and it is easy to set up an incompatible metric. Our results highlight the need for identifying the correct bias metric for each domain, and judging both the data and the model by that metric.

\subsection{Ethical Considerations}
Our study attempts to measure a social issue with technical tools, and therefore it relies on some shortcut heuristics and simplifications that we attempt to make explicit here. In studying gender disparities, we rely on pronouns and only focus on he/him and she/her since our metrics are set up as subtractions. This simplification is not meant to reinforce the gender binary, and we acknowledge that multiple instances of log probability gap can be used as well. The list of identity groups for whom we measure sentiment and downstream classification disparities is taken from~\citet{dixon2018measuring}. This list has been designed for broad coverage, and is not necessarily grounded in any harms that have been experienced by these groups.



\bibliography{acl2020}
\clearpage

\end{document}